\pdfoutput=1

\documentclass[11pt]{article}

\usepackage{emnlp2021}

\usepackage{times}
\usepackage{latexsym}

\usepackage[T1]{fontenc}

\usepackage[utf8]{inputenc}

\usepackage{microtype}

\usepackage{graphicx}
\usepackage{booktabs}
\usepackage{multicol}
\usepackage{multirow}
\usepackage{amsmath}
\usepackage{amssymb}
\usepackage{bbding}
\usepackage{subfigure}
\usepackage{colortbl}
\usepackage{pifont}
\usepackage{CJKutf8}

\newcommand{\ourmethod}{\textsc{CroP}}
\title{\ourmethod{}: Zero-shot Cross-lingual Named Entity Recognition with \\ Multilingual Labeled Sequence Translation}

\author{
  Jian Yang\textsuperscript{\rm 1 \thanks{\ Contribution during internship at Microsoft Research Asia.}}, 
  Shaohan Huang\textsuperscript{\rm 2},
  Shuming Ma\textsuperscript{\rm 2}, 
  Yuwei Yin\textsuperscript{\rm 3}, 
  \\ {\bf Li Dong}\textsuperscript{\rm 2}, 
  {\bf Dongdong Zhang}\textsuperscript{\rm 2}, 
  {\bf Hongcheng Guo}\textsuperscript{\rm 1}, 
  {\bf Zhoujun Li}\textsuperscript{\rm 1 \thanks{\ Corresponding author.}}, 
  {\bf Furu Wei}\textsuperscript{\rm 2} \\
  \textsuperscript{\rm 1}State Key Lab of Software Development Environment, Beihang University \\ 
  \textsuperscript{\rm 2}Microsoft Research Asia; \textsuperscript{\rm 3}The University of Hong Kong \\
  \{jiaya, hongchengguo, lizj\}@buaa.edu.cn; \\ \{shaohanh, shumma, lidong1, dozhang, fuwei\}@microsoft.com; yuweiyin@hku.hk
}

\begin{document}
\maketitle
\begin{CJK*}{UTF8}{gbsn}
\begin{abstract}
Named entity recognition (NER) suffers from the scarcity of annotated training data, especially for low-resource languages without labeled data. Cross-lingual NER has been proposed to alleviate this issue by transferring knowledge from high-resource languages to low-resource languages via aligned cross-lingual representations or machine translation results. However, the performance of cross-lingual NER methods is severely affected by the unsatisfactory quality of translation or label projection. To address these problems, we propose a \textbf{Cro}ss-lingual Entity \textbf{P}rojection framework (\ourmethod{}) to enable zero-shot cross-lingual NER with the help of a multilingual labeled sequence translation model. Specifically, the target sequence is first translated into the source language and then tagged by a source NER model. We further adopt a labeled sequence translation model to project the tagged sequence back to the target language and label the target raw sentence. Ultimately, the whole pipeline is integrated into an end-to-end model by the way of self-training. Experimental results on two benchmarks demonstrate that our method substantially outperforms the previous strong baseline by a large margin of +3$\sim$7 F1 scores and achieves state-of-the-art performance.
\end{abstract}

\section{Introduction}
Named entity recognition (NER) focuses on recognizing entities from raw text into predefined types \cite{conll_2002_ner,conll_2003_ner,survey_ner,teb_ner,adversarial_ner,paraphrasing_ner}, which is an essential component for downstream natural language processing (NLP) tasks, such as information retrieval \cite{information_retrieval_ner} and question answering \cite{question_answering_ner2,question_answering_ner}. However, most of the existing approaches are highly dependent on the annotated training data and do not perform well in low-resource languages. 

Zero-shot cross-lingual NER aims to address this challenging problem by transferring knowledge from the high-resource source language with lots amounts of annotated corpora to those languages without any labeled data~\cite{minimal_resources_ner}. Some methods leverage the cross-lingual representations \cite{weakly_supervised_xner}, where the NER model is trained on the labeled corpus of the source language and then directly evaluated on target languages. Due to the success of multilingual pretrained language models \cite{mbert,xlmr}, these model-based transfer methods have shown a significant improvement in cross-lingual NER. Another line of research is the data-based transfer \cite{unitrans}, which adopts word-to-word translation to project the cross-lingual NER labels. For example, \citet{mulda} employ a multilingual translation model with placeholders for label projection. Nevertheless, these methods are still limited by weak entity projection and do not leverage the unlabeled corpora in target languages. 

\begin{figure}[t]
\begin{center}
    \includegraphics[width=1.0\columnwidth]{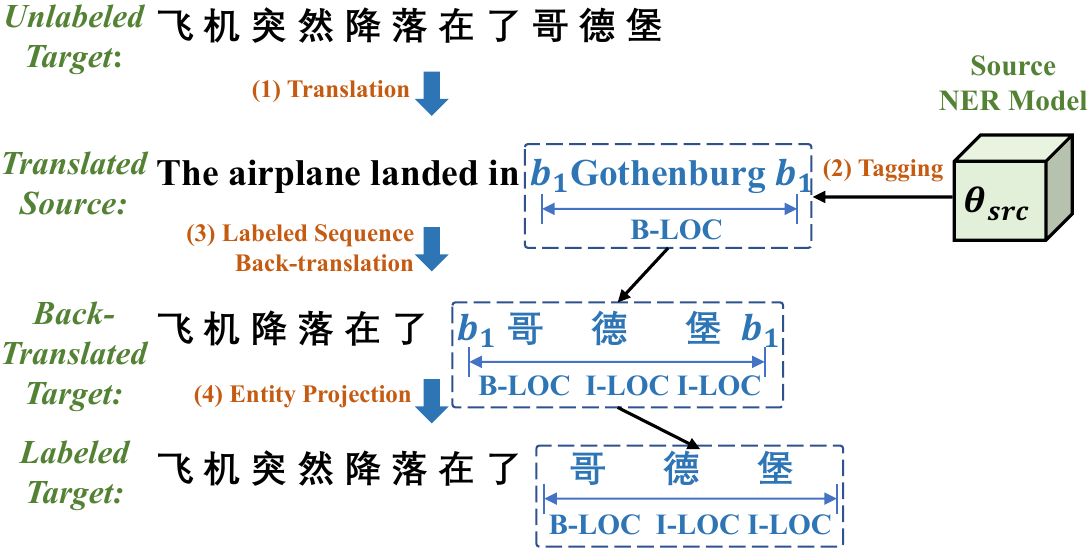}
    \caption{Illustration of our method. It enables cross-lingual zero-shot transfer from source (English) to target (Chinese) language via labeled sequence translation and then entity projection.}
    \label{intro}
\end{center}
\end{figure}

Along the line of using the multilingual model to encourage knowledge transfer among different languages, we propose a \textbf{Cro}ss-lingual Entity \textbf{P}rojection (\ourmethod{}) framework to leverage the unlabeled corpora of target languages, which is supported by a strong multilingual labeled sequence translation model guided by multiple bilingual corpora and the corresponding phrase-level alignment information. 
In Figure \ref{intro}, the unlabeled target sentence is forward translated to the source language and tagged by the source NER model. Then, we use the labeled sequence translation model to back-translate the annotated sentence to the target language. Given the target annotated sentence, we project the entity labels of ``Gothenburg'' to the target raw sentence through lexical matching. Finally, we use self-training to integrate the pipeline into an end-to-end NER model.

Specifically, we construct multilingual corpora to train the labeled sequence translation model, where the aligned spans of the sentence pair are both surrounded by the boundary symbols. We conduct experiments on two benchmarks, including XTREME-40 of 40 languages and CoNLL-5 of 5 languages. Experimental results show that our method reaches new state-of-the-art results. Furthermore, we also evaluate the performance of the multilingual labeled sequence translation model and visualize multilingual sentence representations. Analytic results demonstrate that our method can transfer knowledge among even distant languages.

\section{Zero-shot Cross-lingual NER}
Given the source NER model $\Theta_{ner}^{src}$ only trained on the source NER dataset and the target raw sentence $x=(x_1,\dots,x_m)$ with $m$ words, the zero-shot cross-lingual NER aims to identify each word of target language to predefined types and then obtains the labels $t=(t_1,\dots,t_m)$.
The problem definition of zero-shot cross-lingual NER is described as:
\begin{MiddleEquation}
\begin{align}
    P(t|x) = \prod_{i=1}^{m}P(t_i|x;\Theta_{ner}^{src})
    \label{problem_definition}
\end{align}
\end{MiddleEquation}where the target raw sentence $x$ and labels $t$ have the same length $m$. $t_i$ is the $i$-th label. The source language has annotated labels but the target corpora have no accessible handcrafted labels. $P(t|x)$ represents the predicted distributions of labels. The source NER model $\Theta_{ner}^{src}$ trained on the source annotated corpus is expected to be evaluated on the target language without any labeled dataset. The previous work \cite{unitrans} propose to unify the model-based transfer and data-based transfer with machine translation to transfer knowledge from the source language to the target language.

\section{\ourmethod{}}
\begin{figure*}[t]
\begin{center}
	\includegraphics[width=1.95\columnwidth]{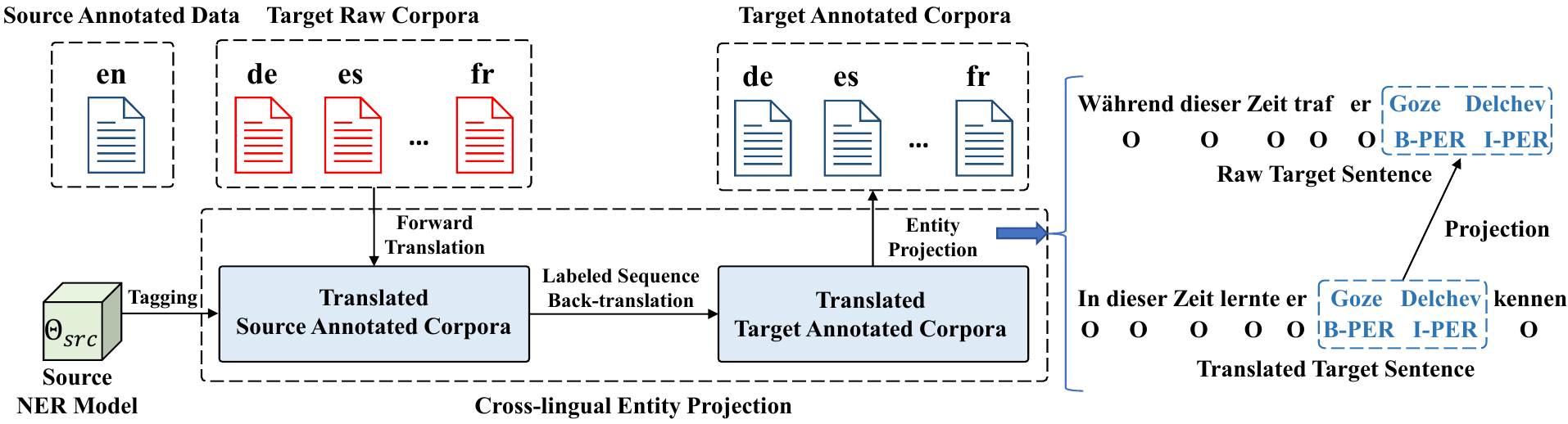}
	\caption{Framework of our proposed method \ourmethod{}, which projects the labels of the translated entities into the target raw data by the multilingual forward translation and labeled sequence back-translation.}
	\label{framework}
\end{center}
\end{figure*}
\subsection{Framework}
In Figure \ref{framework}, given $K$ target languages $L_{tgt}=\{L_k\}_{k=1}^{K}$, the NER model is first trained on the NER dataset $D_{x,t}^{L_{src}}=\{(x^{(i)},t^{(i)})\}_{i=1}^{N}$ of the source language $L_{src}$ with $N$ samples, where $x^{(i)}$ is the input sentence and $t^{(i)}$ contains labels. The raw sentences in $\{D^{L_k}_{x}\}_{k=1}^{K}$ are translated to the source language and tagged by the source NER model. Then, the source annotated corpora are back-translated to the target sentences via a labeled sequence translation model. The labels of the target translated sentences are projected to the target raw corpora to construct the annotated corpora $\{D^{L_k}_{x,f(x)}\}_{k=1}^{K}$, where $f(x)$ is projected label of the sentence $x$ by a simple lexical matching between translated entities and original words. The source corpus $D_{x,t}^{L_{src}}$ and target annotated corpora $\{D^{L_k}_{x,f(x)}\}_{k=1}^{K}$ are further utilized by self-training.

\subsection{Backbone Model for NER}
Our backbone model for NER is comprised of an encoder and a linear classifier to identify entities to predefined types. Given the input sentence $x=(x_1,\dots,x_m)$ with $m$ words, we use the encoder $\Theta_{e}$ to extract top-layer features:
\begin{MiddleEquation}
\begin{align}
    \begin{split}
    H = \text{Encoder}(x; \Theta_{e})
    \label{extractor}
    \end{split}
\end{align}
\end{MiddleEquation}where $H=(h_1,\dots,h_m)$ are the representations of the last encoder layer, where $h_i$
is the $i$-th word representation of the input sentence $x$. $\Theta_{e}$ are parameters of the feature extractor.  

Then, a sequence of representations $H=(h_1,\dots,h_m)$ are fed into a linear classifier with the softmax function to generate the probability distribution of each input word:
\begin{MiddleEquation}
\begin{align}
    \begin{split}
    P(t|x)=\text{Softmax}(W_{c}H+b_{c})
    \label{classifier}
    \end{split}
\end{align}
\end{MiddleEquation}where $t=(t_1,\dots,t_m)$ are corresponding labels of the input sentence, and $\Theta_{ner}=\{W_{c}, b_{c}\}$ represent model parameters of the NER backbone model. $P(t|x) \in R^{m \times T}$ is the predicted probabilities and $T$ is the number of the predefined types. In this work, we set $T=7$ on the XTREME benchmark and $T=9$ on the CoNLL benchmark.

\subsection{Labeled Sequence Translation}
\label{labeled_sequence_translation}
We adopt the multilingual labeled sequence translation (LST) to transfer knowledge from high-resource to low-resource languages. The bilingual pair $x=(x_1,\dots,x_m)$ with $m$ words and $y=(y_1,\dots,y_n)$ with $n$ words are used to construct the pseudo labeled pair $x_p=(x_1,\dots,b_{i},x_{u_{1}:u_{2}},b_{i},\dots,x_m)$ and $y_p=(y_1,\dots,b_{i},y_{v_{1}:v_{2}},b_{i},\dots,y_n)$, where $y_{v_{1}:v_{2}}$ is the target translation of source piece $x_{u_{1}:u_{2}}$. $x_{u_{1}:u_{2}}$ denotes the source phrase from the $u_{1}$-th token to the $u_{2}$-th token and $y_{v_{1}:v_{2}}$ denotes the target phrase from the $v_{1}$-th token to the $v_{2}$-th token. $b_{i}$ is the boundary symbol to indicate the $i$-th entity. We use the alignment tool \texttt{eflomal}\footnote{\url{https://github.com/robertostling/eflomal}} to extract the aligned phrases of the sentence pair. $x_p$ and $y_p$ are used to help the model tackle labeled sequence translation, where $x_p$ and $y_p$ have multiple aligned spans surrounded by boundary tokens. For each sentence pair, we randomly sample 0$\sim$10 aligned spans from the pair and use the boundary symbols \{$b_1,\dots,b_{10}$\} to construct the labeled sequence $x_p$ and $y_p$ in the labeled sequence translation training.

Given bilingual corpora $D_{b}=\{D_{b}^{L_k}\}_{k=1}^{K}$ of $K$ languages, where one side is the source language $L_{src}$ and the other side is the language $L_{k} \in L_{tgt}$, the multilingual model is trained on corpora $D_{b}$:
\begin{SmallEquation}
\begin{align}
    \begin{split}
    \mathcal{L}_{t} =-\sum_{k=1}^{K} \mathbb{E}_{x,y \in D_{b}^{L_{k}}} \left[ \log P(y|x;\Theta_{mt}) \right] 
    \label{mt}
\end{split}
\end{align}
\end{SmallEquation}where $\Theta_{mt}$ are parameters of translation model.

To support labeled sequence translation (LST), we use the sentence pair to construct training samples, where the aligned spans in the sentence pair are surrounded by the boundary symbols using phrase-level alignment pairs. In Figure \ref{translation}, $x$ and $y$ are sentence pair. The aligned fragments of the source sentence and target sentence are both annotated by the boundary symbols. These samples are used for the training of labeled sequence translation:
\begin{SmallEquation}
\begin{align}
    \begin{split}
    \mathcal{L}_{lst} =-\sum_{k=1}^{K} \mathbb{E}_{x,y \in D_{b}^{L_{i}}} \left[ \log P(y_p|x_p;\Theta_{mt}) \right] 
    \label{lt}
\end{split}
\end{align}
\end{SmallEquation}where $(x_p, y_p)$ is the sentence pair constructed by the original sentence pair and the phrase-level alignment pairs.

Our model is optimized by jointly minimizing the translation objective and labeled sequence translation objective:
\begin{SmallEquation}
\begin{align}
\begin{split}
    \mathcal{L}_{mt}= \alpha \mathcal{L}_{t}  + (1-\alpha)\mathcal{L}_{lst}
    \label{mt_all}
\end{split}
\end{align}
\end{SmallEquation}where $\mathcal{L}_{t}$ is the objective of multilingual translation and $\mathcal{L}_{lst}$ is the objective of the multilingual labeled sequence translation. We alternate two training objectives by setting $\alpha=0.5$. Our multilingual model supports (i) multilingual translation and (ii) labeled sequence translation. After alternately training on two objectives, we obtain the final multilingual translation model $\Theta_{mt}$. Once the multilingual training is done, our model serves as the off-the-shelf multilingual labeled translation model and does not require alignments.

During the inference stage, the source sentence $x$ with labels is switched to labeled sequence $x_p$, where all entities are surrounded by indicators. Then, the model translates the source labeled sentence $x_p$ to the target labeled sentence $y_p$. The boundary symbol indicates the entities in the translation sentence. For example, the translation phrase $y_{v_{1}:v_{2}}$ have the same NER labels with the source phrase $x_{u_{1}:u_{2}}$, where both phrases are surrounded by the boundary token $b_{i}$.

\subsection{Cross-lingual Entity Projection}
\label{crosslingual_entity_projection}
Given the labeled corpus $D_{x,t}^{L_{src}}=\{(x^{(i)},t^{(i)})\}_{i=1}^{N}$ of the source language $L_{src}$ and the unlabeled corpora $\{D^{L_k}_{x}\}_{k=1}^{K}$ of $K$ languages, the source NER model $\Theta_{ner}^{src}$ is used to tag the unlabeled training corpora of target languages, aided by the labeled translation model $\Theta_{mt}$.

\paragraph{Forward Translation} The target raw corpora $\{D^{L_k}_{x}\}_{k=1}^{K}$ of $K$ languages are translated into the source language and tagged by the source NER model $\Theta_{ner}^{src}$. We obtain the source labeled translated corpora $\{D^{L_{k}^{src}}_{x,f(x)}\}_{k=1}^{K}$, where $f(\cdot)$ is the predictor of the source NER model $\Theta_{ner}^{src}$.

\begin{figure}[t]
\begin{center}
	\includegraphics[width=1.0\columnwidth]{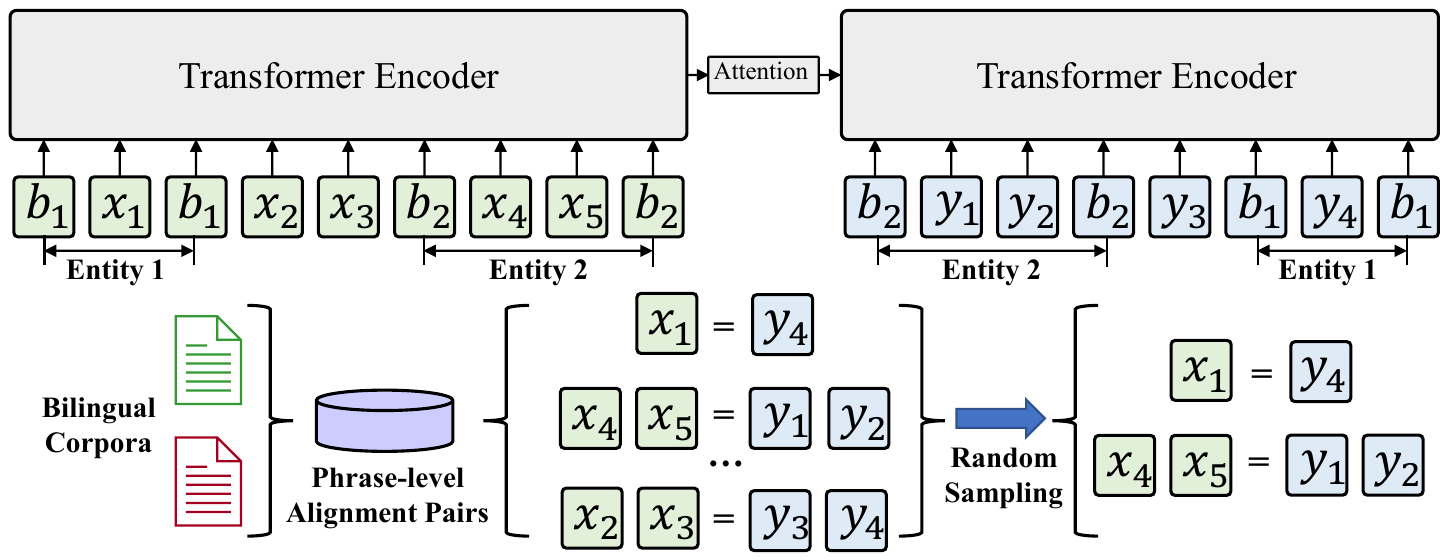}
	\caption{Multilingual labeled sequence translation.}
	\label{translation}
\end{center}
\end{figure}

\paragraph{Labeled Sequence Back-translation} The source annotated corpora are back-translated to the target languages with entity labels. In Figure \ref{translation}, the source sentence $x_p=(b_1,e_1,b_1,x_2,x_3,b_2,e_2,b_2)$ is translated into the target sentence $y_p=(b_2,e_2,b_2,y_3,b_1,e_1,b_1)$. The boundary symbols $b_1$ and $b_2$ are used to locate the translated entities $e_1$ and $e_2$ in $y_p$. We obtain the back-translated data $\{D^{L_{k}^{bt}}_{x,f(x)}\}_{k=1}^{K}$ by the translation model $\Theta_{mt}$.

\paragraph{Entity Matching} Given the target translated entities with labels $D_{tgt}$, we search the matched entities in the unlabeled target sentence by lexical matching (string matching word by word). In Figure \ref{intro}, ``哥德堡'' in the unlabeled sentence matches ``哥德堡'' in the translated sentence, so ``哥德堡'' is labeled with the same entity type \texttt{LOC} (Location).
The labels of translated entities are projected into the raw sentence to construct target labeled corpora $\{D^{L_{k}}_{x,f(x)}\}_{k=1}^{K}$. Finally, the target annotated corpora and the original corpus $D_{x,t}^{L_{src}}$ are used for multilingual NER model training.

\subsection{Self-training}
Given a labeled corpus $D^{L_{src}}_{x,t}$ of the source language and target unlabeled corpora $\{D^{L_k}_{x}\}_{k=1}^{K}$ of target languages, the training objective based on $D^{L_{src}}_{x,t}$ is formulated as below:
\begin{SmallEquation}
\begin{align}
\begin{split}
    \mathcal{L}_{src}=\mathbb{E}_{x,t \in D_{x,t}^{L_{src}}} \left[-\log P(t|x;\Theta_{ner}^{all}) \right]
    \label{original_ner_loss}
\end{split}
\end{align}
\end{SmallEquation}where $\Theta_{ner}^{all}$ are NER model parameters.

Then, we leverage the source NER model $\Theta_{src}^{ner}$ trained on the labeled corpus to project the entity labels to the target raw corpora described in Section \ref{crosslingual_entity_projection} and get labeled corpora $\{D^{L_k}_{x,f(x)}\}_{k=1}^{K}$. The multilingual corpora of target languages with predicted labels are adopted to train a neural network with the combined loss function $\mathcal{L}_{tgt}$ as below:
\begin{SmallEquation}
\begin{align}
\begin{split}
    \mathcal{L}_{tgt}=\sum_{L_{k} \in L_{tgt}} \mathbb{E}_{x,y \in D_{x,f(x)}^{L_{k}}}\left[-\log P(f(x)|x;\Theta_{ner}^{all}) \right]
    \label{pseudo_ner_loss}
\end{split}
\end{align}
\end{SmallEquation}where $x$ is the golden target data and $f(x)$ is the pseudo label generated by the cross-lingual entity projection.

The multilingual NER model is jointly trained on the original dataset and target corpora with labels:
\begin{SmallEquation}
\begin{align}
\begin{split}
    \mathcal{L}_{all}=\mathcal{L}_{src} + \mathcal{L}_{tgt}
    \label{all_ner_loss}
\end{split}
\end{align}
\end{SmallEquation}where $\mathcal{L}_{src}$ and $\mathcal{L}_{tgt}$ are training objectives on the original and distilled dataset.

\section{Experiments}
\subsection{Dataset}

\paragraph{CCaligned} Our labeled multilingual model continues to be tuned on the same training data called CCaligned \cite{ccaligned} as the previous work \cite{m2m,flores}. We use a collection of parallel data in different languages from the CCaligned dataset, where the parallel data is paired with English and other 39 languages. The valid and test sets are from the FLORES-101 dataset \cite{flores}. 

\paragraph{CoNLL-5} Following the previous work \cite{unitrans}, we construct a cross-lingual dataset from CoNLL-2002 \cite{conll_2002_ner} for Spanish (es) and Dutch (nl) NER, CoNLL-2003 \cite{conll_2003_ner} for English (en) and German (de)
NER, and NoDaLiDa-2019 \cite{norwegian9_ner} for Norwegian
(no) NER. All entities are classified into 4 entity types in \textit{BOI-2} format, including \texttt{LOC}, \texttt{MISC}, \texttt{ORG}, and \texttt{PER}. Each dataset is split into training, dev, and test set. Detailed statistics can be found in Table \ref{details_conll_5}.

\paragraph{XTREME-40} The proposed method is further evaluated on the cross-lingual NER dataset from the XTREME benchmark \cite{xtreme}. Named entities in Wikipedia are annotated with \texttt{LOC}, \texttt{PER}, and \texttt{ORG} tags in \textit{BOI-2} format. Following the previous work \cite{xtreme}, we use the same split for the training, dev, and test set.

\paragraph{Post-Processing}
The synthetic data is post-processed to train the multilingual NER model.
(i) We use the language detection toolkit\footnote{\url{https://github.com/saffsd/langid.py}} to filter the translated sentence with the incorrect language.
(ii) We delete sequences, which exceed the maximum length (128 words) and only contain $O$ (other) tags.
(iii) The NER model trained on the multilingual corpora is directly employed to tag the unlabeled corpora. The discarded sentence is re-labeled by the multilingual NER model.
Finally, we combine the labels predicted by the source NER model $\Theta_{ner}^{src}$ trained on the original dataset and the multilingual NER model $\Theta_{ner}^{all}$ trained by self-training to improve the accuracy of label projection.

\subsection{Baselines and Evaluation}
Our method is compared with the different baselines initialized by cross-lingual pretrained models including \textbf{mBERT} \cite{mbert} and \textbf{XLM-R} \cite{xlmr} for model-based transfer. We also conduct experiments without any pretrained model on the \textbf{Transformer} \cite{transformer} architecture. \textbf{UniTrans} \cite{unitrans} unifies both model transfer and data transfer for cross-lingual NER. Following this line of research, \textbf{MulDA} \cite{mulda} proposes a sequence translation method to translate labeled training data of the source languages to other languages and avoids the word order change caused by word-to-word or phrase-to-phrase translation. Besides, we also produce the results of \textbf{Translate-Train}, where the labeled source corpus is translated into the other labeled corpora of multiple languages using our multilingual model. 
Following the previous work \cite{conll_2002_ner,unitrans}, the metrics are the entity-level precision, recall, and F1 scores. For simplicity, we report the F1 scores of different methods in all tables.

\subsection{Training Details}
\paragraph{Multilingual Labeled Translation}
The pretrained multilingual model M2M$_{large}$\footnote{\url{https://dl.fbaipublicfiles.com/flores101/pretrained_models/flores101_mm100_615M.tar.gz}} is adopted as the translation model, which has 12 layers with an embedding size of 1024 and 16 attention heads. We continue fine-tuning the model with Adam ($\beta_{1}=0.9$, $\beta_{2}=0.98$) on the labeled corpora constructed by the multilingual corpora and the alignment pairs from CCAligned\footnote{\url{https://opus.nlpl.eu/CCAligned.php}}, where the parallel data is paired with English and other 39 languages and the phrase-level alignment pairs are extracted by the alignment tool\texttt{eflomal}. The learning rate is set as 1e-4 with a warm-up step of 4,000. The batch size is set as 1536 tokens on 32 A100 GPUs.

\paragraph{Cross-lingual NER} For a fair comparison, we implement all methods using the same architecture and model size. We separately adopt the base architecture of Transformer, mBERT, and XLM-R as the backbone model, which all have 12 layers with an embedding size of 768, a feed-forward network size of 3072, and 12 attention heads. We set the batch size as 24 for CoNLL-5 and 32 for XTREME-40. The NER model is trained on CoNLL-5 for 15 epochs and XTREME-40 for 10 epochs, where the warm-up step is the 10\% steps of the whole training steps. The synthetic data is post-processed to train the multilingual NER model. We delete sequences, which exceed the maximum length (128 words) and only contain $O$ (other) tags. The NER model trained on the multilingual corpora is directly employed to tag the unlabeled corpora. The discarded sentence is re-labeled by the multilingual NER model. Finally, we combine the labels predicted by the source NER model $\Theta_{ner}^{src}$ trained on the original dataset and the multilingual NER model $\Theta_{ner}^{all}$ trained by self-training in Equation \ref{all_ner_loss} to improve the accuracy of label projection.

\subsection{Main Results}

\paragraph{CoNLL-5} Table \ref{conll_5} presents the results of our method and previous baselines on transferring knowledge from English to other four languages, including es, nl, de, no. We can observe that the XLM-R gains strong improvement compared to previous baselines due to the effective cross-lingual transfer.
Based on the cross-lingual pretrained model, our method can leverage cross-lingual entity projection to further encourage transferability from the NER model of the source language to the multilingual NER model of all languages. Our method significantly outperforms the previous strong baseline UniTrans on average, especially on German by a large margin +5.3 points. It can be attributed to our multilingual model, which has better translation quality on German and Norwegian than Spanish and Dutch.

\paragraph{XTREME-40} 
Table \ref{xtreme_40} compares the performance of our method with previous relevant methods initialed by different cross-lingual pretrained language models including mBERT and XLM-R. Given our translation model, the multilingual translated annotated corpora (\textbf{Translate-Train}) from the data of source languages can be used to improve the model performance compared to the XLM-R. Particularly, our proposed method gains significant improvement compared to other languages by a large margin (nearly +6 F1 points), due to the effectiveness of cross-lingual entity projection. All experimental results demonstrate that our proposed framework strengthens transferability from the source language to nearly 39 target languages.

\begin{table}[t]
    \centering
    \resizebox{1.0\columnwidth}{!}{
    \begin{tabular}{c|c|c|c|c}
        \toprule
        Language & Type &Train &Dev &Test  \\ \midrule
        English (en) & \#Sentences &15.0K &3.5K &3.7K \\
        (CoNLL-2003) &  \#Entities &23.5K &6.0K &5.7K \\
        \midrule
        German (de)  & \#Sentences &12,7K &3.1K &3.2K \\
        (CoNLL-2003) & \#Entities  &11.9K &4.8K &3.7K \\
        \midrule
        Spanish [es] & \#Sentences &8.3K  &1.9K &1.5K \\
        (CoNLL-2002) & \#Entities   &18.8K &4.3K &3.6K \\
        \midrule
        Dutch [nl]  & \#Sentences   &15.8K &2.9K &5,2K \\
        (CoNLL-2002)& \#Entities   &13.3K &2.6K &3,9K \\
        \midrule
        Norwegian [no] & \#Sentences &15.7K &2.4K &1.9K \\
        (NoDaLiDa-2019)& \#Entities  &10.9K &1.6K &1.4K \\
        \bottomrule
    \end{tabular}}
    \caption{Statistics of the CoNLL-5 \cite{conll_2002_ner,conll_2003_ner} and NoDaLiDa \cite{norwegian9_ner} NER benchmarks.}
    \label{details_conll_5}
\end{table}

\begin{table}[t]
    \centering  
    \resizebox{1.0\columnwidth}{!}{
    \begin{tabular}{l|ccccc}
        \toprule
        & es & nl & de & no & Average \\ \midrule
        \citet{cross_lingual_word_clusters_unitrans_data}$^{\dag}$     & 59.3 & 58.4 & 40.4 & - & - \\
        \citet{wikification_unitrans_data}$^{\dag}$      & 60.6  & 61.6  & 48.1 & - & - \\
        \citet{word_vectors_unitrans_data}$^{\dag}$      & 65.1  & 65.4  & 58.5 & - & -\\ 
        \citet{cheap_translation_unitrans_data}$^{\dag}$ & 64.1  & 63.4  & 57.2 & - & -\\ 
        \citet{minimal_resources_ner}$^{\dag}$           & 72.4  & 71.3  & 57.8 & - & -\\ 
        \citet{zero_resource_unitrans_data}$^{\dag}$     & 73.5  & 69.9  & 61.5  & - & -\\ 
        \citet{projection_mt_unitrans_data}$^{\dag}$     & 75.9  & 74.6  & 65.2  & - & -\\ 
        \citet{bert_unitrans_data}$^{\dag}$              & 74.5  & 79.5  & 71.1  & - & -\\ 
        \citet{meta_learning_unitrans_data}$^{\dag}$     & 76.8  & 80.4  & 73.2 & - & -\\ 
        \midrule
        mBERT \cite{mbert}                & 74.6 & 77.9  & 75.0 & 77.4 & 76.2  \\
        XLM-R \cite{xlmr}                 & 77.4 & 78.9  & 73.4 & 80.9 & 77.7  \\
        \quad+Translate-Train             & 77.8 & 79.2  & 74.2 & 81.3 & 78.1  \\ 
        UniTrans$^{\dag}$ \cite{unitrans} & \textbf{79.3} & \textbf{82.9}  & 74.8 & 81.2 & 79.6  \\
        MulDA \cite{mulda}                       & 77.5 & 78.4  & 78.2 & 82.1 & 79.1  \\ 
        \textbf{\ourmethod{} (Our Method)}       & 78.1 & 79.5  & \textbf{80.1} & \textbf{83.1}  & \textbf{80.2}  \\
        \bottomrule
    \end{tabular}
    }
    \caption{Results of our proposed method \ourmethod{} and prior state-of-the-art methods for zero-resource cross-lingual NER. The dag symbol represents that the score is directly reported from the previous work.
    }
    \label{conll_5}
\end{table}

\begin{table*}[t]
\centering
\resizebox{1.0\textwidth}{!}{
\begin{tabular}{l|cccccccccccccccccccc}
\toprule
\multicolumn{21}{c}{\textit{Initialized By Pretrained Cross-lingual Language Model mBERT}} \\
\midrule
\textbf{Method}                    & af   & ar   & bg   & bn   & de   & el   & es   & et   & eu   & fa   & fi   & fr   & he   & hi   & hu   & id   & it   & ja   & jv   & ka \\ \midrule
mBERT \cite{mbert}                 & 76.9 & 44.5 & 77.1 & 68.8 & 78.8 & 71.6 & 74.0 & 76.3 & 68.0 & 48.2 & 77.2 & 79.7 & 56.5 & 66.9 & 76.0 & 46.3 & \textbf{81.1} & 28.9 & \textbf{66.4} & 67.7 \\
\quad+Translate Train              & 74.5 & 37.6 & 77.8 & 73.2 & 77.2 & 74.9 & 69.4 & 74.1 & 63.2 & 43.1 & 75.9 & 76.1 & 55.4 & 68.1 & 77.2 & \textbf{48.2} & 77.2 & 36.6 & 55.1 & 64.4 \\
UniTrans \cite{unitrans}           & 78.2 & 47.0 & 79.5 & 74.6 & 79.8 & 75.6 & 75.2 & 76.5 & 67.2 & 49.3 & 75.6 & 80.1 & 58.4 & 72.1 & 77.9 & 44.6 & 78.3 & 37.6 & 56.2 & 69.9 \\
\textbf{\ourmethod{} (Our Method)} & \textbf{81.0} & \textbf{48.0} & \textbf{80.8} & \textbf{74.9} & \textbf{80.3} & \textbf{78.7} & \textbf{84.2} & \textbf{78.3} & \textbf{70.6} & \textbf{63.2} & \textbf{79.1} & \textbf{83.5} & \textbf{64.7} & \textbf{77.1} & \textbf{82.5} & 46.4 & 79.9 & \textbf{45.3} & 57.7 & \textbf{74.1} \\ \midrule\midrule
\textbf{Method}                    & kk   & ko   & ml   & mr   & ms   & my   & nl   & pt   & ru   & sw   & ta   & te   & th   & tl   & tr   & ur   & vi   & yo   & zh   & \textbf{Avg$_{all}$}   \\ \midrule
mBERT \cite{mbert}                 & 50.4 & 60.2 & 53.7 & 56.2 & 61.9 & 47.6 & 82.1 & 79.6 & 65.2 & 72.8 & 50.8 & 46.8 & 0.4  & 71.2 & 75.5 & 36.9 & 69.7 & 51.7 & 44.1 & 61.7  \\
\quad+Translate Train              & 48.2 & 61.2 & 61.0 & 58.7 & 67.5 & 57.3 & 79.6 & 78.4 & 61.2 & \textbf{69.2} & 62.7 & 51.2 & 2.4  & 72.7 & 72.6 & 58.9 & 69.5 & 51.1 & 45.3 & 62.3  \\
UniTrans \cite{unitrans}           & 52.5 & 61.4 & 63.5 & 62.3 & 65.8 & 59.2 & 82.4 & 80.3 & 64.8 & 65.2 & 63.2 & 56.1 & 3.1  & 73.4 & 77.9 & 64.1 & 69.7 & 50.1 & 47.4 & 64.5  \\
\textbf{{\ourmethod} (Our Method)} & \textbf{54.9} & \textbf{62.6} & \textbf{72.7} & \textbf{70.6} & \textbf{71.1} & \textbf{61.3} & \textbf{84.6} & \textbf{81.7} & \textbf{69.7} & 68.3 & \textbf{64.9} & \textbf{61.6} & \textbf{3.9}  & \textbf{76.9} & \textbf{80.4} & \textbf{78.0} & \textbf{70.0} & \textbf{51.8} & \textbf{54.4} & \textbf{68.4}  \\
\midrule
\multicolumn{21}{c}{\textit{Initialized By Pretrained Cross-lingual Language Model XLM-R}} \\
\midrule
\textbf{Method}                   & af   & ar   & bg   & bn   & de   & el   & es   & et   & eu   & fa   & fi   & fr   & he   & hi   & hu   & id   & it   & ja   & jv   & ka \\ \midrule
XLM-R \cite{xlmr}                 & 74.6 & 46.0 & 78.0 & 68.3 & 75.2 & 75.7 & 70.2 & 72.2 & 59.9 & 52.0 & 75.8 & 76.6 & 52.4 & 69.6 & 78.2 & 47.4 & 77.7 & 21.0 & 61.8 & 66.5 \\
\quad+Translate Train             & 76.2 & 47.8 & 79.2 & 74.3 & 75.8 & 67.7 & 68.4 & 75.8 & 61.2 & 41.0 & 76.8 & 76.4 & 55.0 & 71.9 & 76.0 & \textbf{50.6} & 78.1 & 35.4 & 54.7 & 68.4 \\
UniTrans \cite{unitrans}          & 78.1 & \textbf{48.1} & 79.3 & 74.6 & 75.2 & 74.9 & 73.8 & 76.9 & 62.7 & 49.2 & 74.6 & 76.5 & 53.4 & 70.4 & 76.9 & 48.6 & 77.3 & 21.6 & 62.2 & 66.8 \\
\textbf{\ourmethod{} (Our Method)} & \textbf{80.3} & 45.2 & \textbf{80.4} & \textbf{75.7} & \textbf{79.6} & \textbf{78.5} & \textbf{83.1} & \textbf{77.2} & \textbf{66.8} & \textbf{65.5} & \textbf{77.9} & \textbf{82.9} & \textbf{63.5} & \textbf{77.4} & \textbf{81.6} & 46.1 & \textbf{78.8} & \textbf{45.4} & \textbf{63.2} & \textbf{74.0} \\ \midrule\midrule
\textbf{Method}                   & kk   & ko   & ml   & mr   & ms   & my   & nl   & pt   & ru   & sw   & ta   & te   & th  & tl   & tr   & ur   & vi   & yo   & zh   & \textbf{Avg$_{all}$} \\ \midrule
XLM-R \cite{xlmr}                 & 43.2 & 49.9 & 62.3 & 59.6 & 67.3 & 53.5 & 80.2 & 78.1 & 64.3 & \textbf{70.3} & 55.0 & 50.1 & 3.0 & 69.4 & 78.1 & 63.6 & 68.2 & 47.5 & 27.7 & 61.3\\
\quad+Translate Train             & 40.1 & 55.5 & 60.0 & 59.8 & 69.8 & 61.6 & 79.6 & 76.4 & 60.9 & 70.0 & 63.7 & 50.7 & 3.4 & 74.7 & 72.3 & 62.7 & 69.6 & 46.8 & 41.2 & 62.3   \\
UniTrans \cite{unitrans}          & 46.5 & 57.2 & 65.5 & 64.5 & 70.2 & 62.6 & 81.8 & 79.4 & 68.8 & 68.9 & 65.1 & 56.1 & 4.8 & 74.8 & 76.4 & 71.0 & 69.8 &\textbf{55.1} & 44.4 & 64.2 \\
\textbf{\ourmethod{} (Our Method)} & \textbf{50.2} & \textbf{59.8} & \textbf{73.8} & \textbf{71.6} & \textbf{71.8} & \textbf{69.0} & \textbf{83.5} & \textbf{81.5} & \textbf{70.2} & 69.0 & \textbf{65.6} & \textbf{59.9} & 3.1 & \textbf{75.5} & \textbf{80.5} & \textbf{80.4} & \textbf{70.1} & 52.6 & \textbf{50.3} & \textbf{68.2}\\
\bottomrule
\end{tabular}}
\caption{Results of our proposed method {\ourmethod} and other relevant baselines for cross-lingual NER. ``Avg$_{all}$'' represents the average F1 scores of all 39 languages on the test set of the XTREME-40 benchmark.}
\label{xtreme_40}
\end{table*}

\paragraph{Ablation Study}  To verify the effectiveness of our method, we separately study the effects of the model-based transfer by cross-lingual pretrained model and the data-based transfer by cross-lingual entity projection. Our method has two advantages: (1) the model is trained on the original multilingual corpora with pseudo labels, which avoids the extra translation error. (2) our method uses the multilingual model trained on 41 languages to improve the entity projection of low-resource languages. In Table \ref{ablation},  Transformer {\large{\ding{174}}} without any transfer methods gets the worst performance (only 15.1 F1 scores). Our method {\large{\ding{173}}} without any pretrained model outperforms Transformer {\large{\ding{174}}} by +43.0 F1 points, which has the similar transferability to the cross-lingual pretrained language models. Combining the merits of the cross-lingual pretrained model and self-training for multiple languages, we obtain the best performance on the XTREME-40 benchmark.
\begin{table}[t]
\centering
\resizebox{0.9\columnwidth}{!}{
\begin{tabular}{l|l|ccccc|c}
\toprule
ID & \textbf{Method}                                          & es   & eu   & ta    & tl   & zh   & Avg$_{all}$     \\  \midrule
{\large{\ding{172}}} & \ourmethod{}              & \textbf{83.1} & \textbf{66.8} & \textbf{65.6}  & \textbf{75.5} & \textbf{50.3} & \textbf{68.2}   \\ 
{\large{\ding{173}}} & {\large{\ding{172}}} - XLM-R           & 72.6 & 62.6 & 55.9  & 72.2 & 38.9 & 58.1   \\
{\large{\ding{174}}} & {\large{\ding{173}}} - Transfer   & 20.0 & 14.7 & 6.0   & 33.7 & 1.6  & 15.1    \\
\bottomrule
\end{tabular}}
\caption{Ablation study of our proposed method. Avg$_{all}$ denotes the average F1 scores of 39 languages.}
\label{ablation}
\vspace{-5pt}
\end{table}

\begin{figure}[t]
\centering
    \subfigure[Gold]{
    \includegraphics[width=0.43\columnwidth]{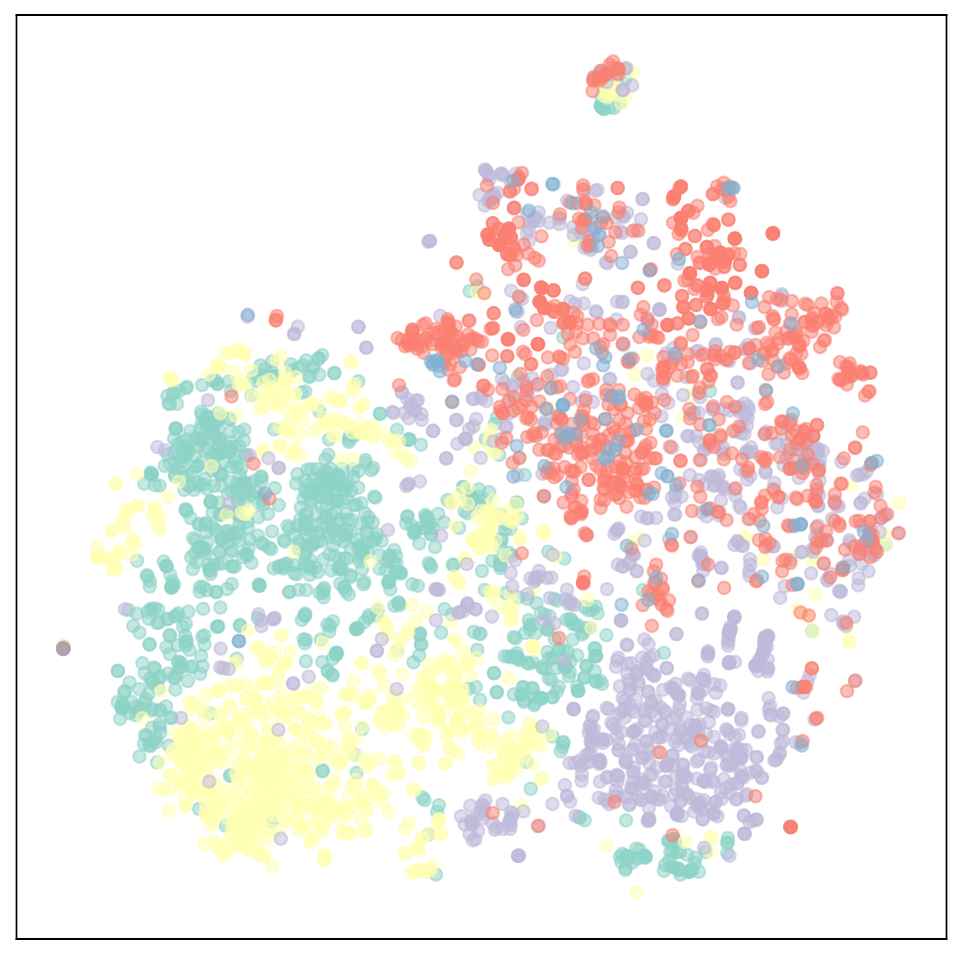}\quad
    \label{tsne_gold}
    }
    \subfigure[Translation]{
    \includegraphics[width=0.43\columnwidth]{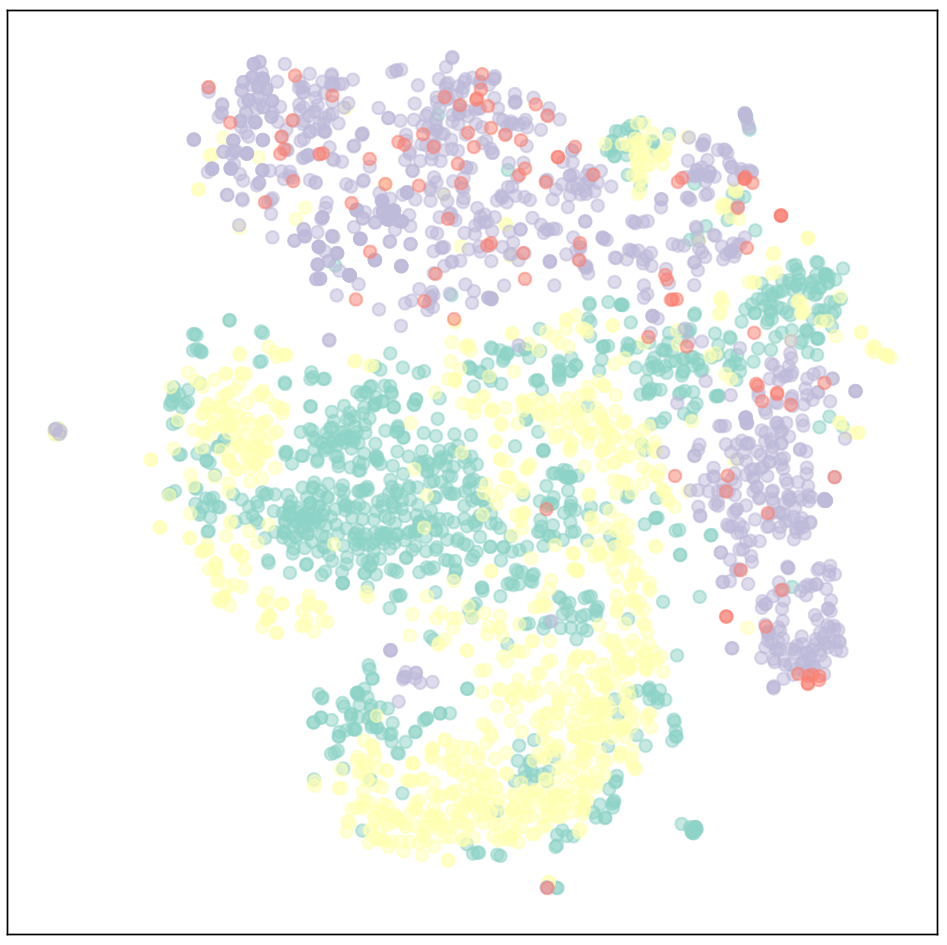}\quad
    \label{tsne_our}
    }
    \caption{t-SNE \cite{t_SNE} visualization of the sentence representations for the golden data (a) and the translated data generated by our multilingual model (b). Each color denotes one language.} 
    \vspace{0pt}
    \label{tsne_figures}
\end{figure}

\paragraph{Distribution of Multilingual Corpora} 
An important difference between our method and the previous baselines is that we provide an effective way to leverage the unlabeled corpora of target languages. The raw data is first translated to the source language data and annotated by the NER model trained on the original dataset. Then, the translated source sentences are back-translated to target languages, where the entity labels are projected to the target raw words. Our cross-lingual entity projection avoids the extra translation errors instead of direct utilization for translated labeled corpora. In Figure \ref{tsne_gold}, we visualize the distribution of the encoder representations by randomly sampling 1K sentences of each language from the target golden corpora. Figure \ref{tsne_our} shows the distribution of the round-trip translated target corpora. We observe that the distribution of translated corpora has changed a lot since there are incorrectly translated words highly affected by translation quality, especially for low-resource languages.

\paragraph{Performance of Multilingual Translation}
To ensure the effectiveness of our method, we evaluate the translation performance of 40 languages between M2M \cite{flores} and our labeled sequence multilingual translation model on the FLORES-101 benchmark. Compared to M2M, our model supports the additional language eu by extending the fine-tuning data. Therefore, we report the SentencePiece-based BLEU using SacreBLEU\footnote{\url{https://github.com/ngoyal2707/sacrebleu}} of 39 translation directions except eu languages.
\begin{table}[t]
\centering
\resizebox{0.95\columnwidth}{!}{
\begin{tabular}{l|cc|c}
\toprule
      & Avg$_{X \to En}$  & Avg$_{En \to X}$   & Avg$_{all}$  \\
\midrule
M2M \cite{flores}            & 24.50 & 22.08  & 23.61  \\
Our Multilingual Model     & \textbf{32.70} & \textbf{30.31}  & \textbf{31.51}  \\
\bottomrule
\end{tabular}}
\caption{Comparison of BLEU points between M2M \cite{flores} and our multilingual model on the FLORES-101 benchmark of 39 languages.}
\label{mulitlingual_transltion_results}
\vspace{-10pt}
\end{table}

\paragraph{Quality of Labeled Sequence Translation}
Section \ref{labeled_sequence_translation} introduces the multilingual labeled sequence translation, where the entities are surrounded by the boundary symbols and then translated to the target language. The multilingual model is trained with the bilingual corpus and the corresponding phrase-level alignment pairs to ensure the quality of labeled sequence translation. We calculate the precision of the baseline model and our model by randomly sampling 250 sentence pairs of each language from the whole training data and human evaluation. More specifically, we check whether the boundary symbol surrounds the equivalent entity in both source and target sentences. The baseline model encounters the problem of boundary symbol missing and incorrect alignment.
In Table \ref{pattern_transltion_results}, our model guided by the phrase-level alignment information outperforms the baseline model showing the strength of our method.

\begin{table}[t]
\centering
\resizebox{0.95\columnwidth}{!}{
\begin{tabular}{l|c|ccc}
\toprule
        & Alignment Pairs                                    & En$\to$Zh  & En$\to$De   & En$\to$Fr  \\
\midrule
\multirow{2}{*}{Our Multilingual Model}   &    & 84.4\%           &    84.8\%       &  86.8\%    \\
                                            &    $\checkmark$ & \textbf{97.2\%}    & \textbf{95.6\%}   & \textbf{94.8\%}    \\
\bottomrule
\end{tabular}}
\caption{Comparison of labeled translation quality between our multilingual model with alignment pairs and the counterpart without alignment information.}
\label{pattern_transltion_results}
\vspace{-10pt}
\end{table}

\begin{figure}[t]
\begin{center}
	\includegraphics[width=0.7\columnwidth]{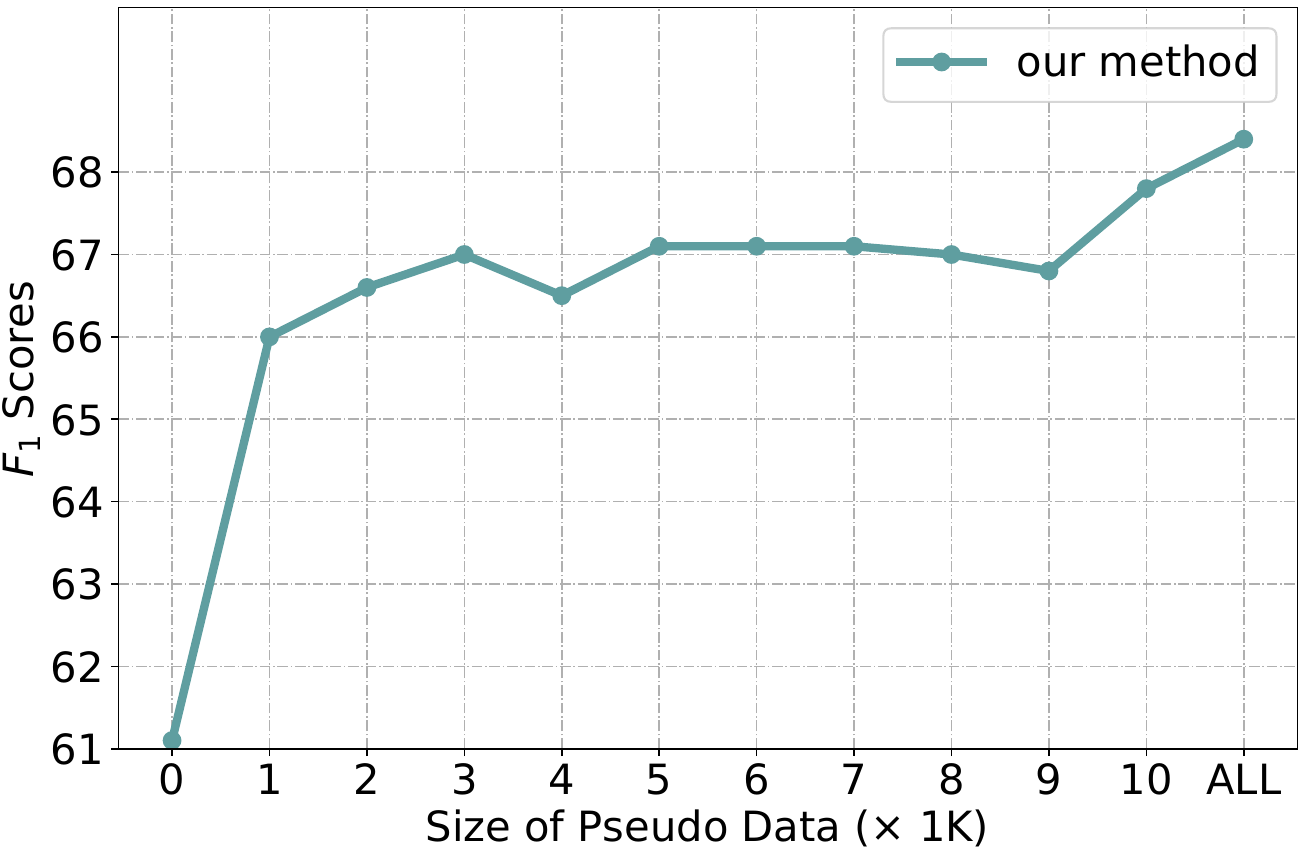}
	\caption{Evaluation results on the original source annotated corpus and pseudo corpora with different training sizes by randomly down-sampling.}
	\label{data_size}
\end{center}
\end{figure}

\paragraph{Effect of Training Data Size}
To discuss the effect of the target labeled corpora, we plot F1 scores with different training data sizes in Figure \ref{data_size}. The performance is influenced by the ratio between the size of the original dataset (20K sentences) and the multilingual corpora (400K sentences after filtering). We randomly sample $N=$ $\{1K, 2K, \dots, ALL\}$ sentences from the whole corpora to train the NER model.
With the training data size increasing, the NER model gets better performance. Surprisingly,  only $1K$ pseudo annotated sentences bring large improvement to the zero-shot cross-lingual NER, which benefits from knowledge transfer of the multilingual self-training. When the size of target annotated corpora is greater than 10K, our method gets exceptional performance.

\begin{figure}[t]
\begin{center}
	\includegraphics[width=1.0\columnwidth]{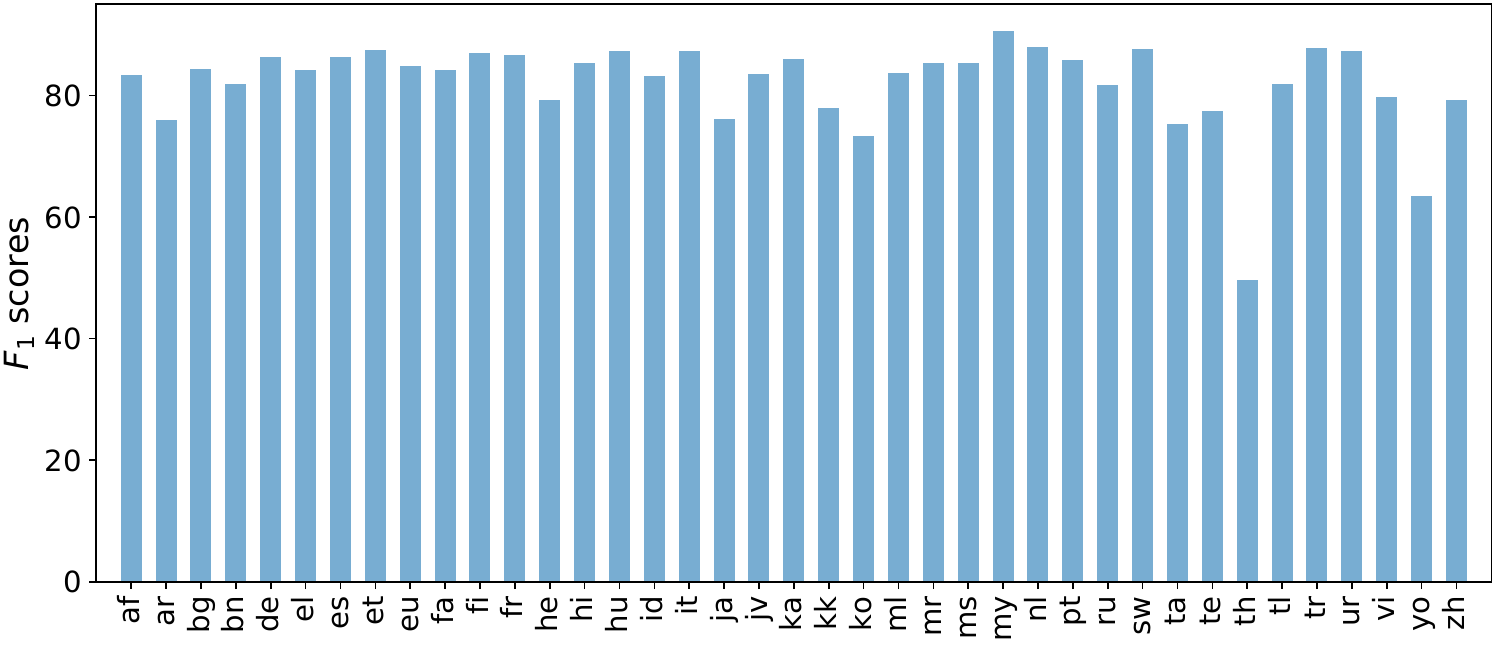}
	\caption{F1 scores of cross-lingual entity projection based on the golden labels. The languages are ordered by alphabet order.}
	\label{pseudo_label_f1}
 	\vspace{-10pt}
\end{center}
\end{figure}

\paragraph{Quality of Entity Projection}
Given the target annotated translated sentence and the raw sentence, our method searches the matched entity and projects the labels to the raw sentence. After filtering the sentences, we utilize the labeled sentences with pseudo labels for multilingual NER model training. Figure \ref{pseudo_label_f1} reports the F1 scores of the projected labels of the target corpora compared to the ground-truth labels, where each language has high F1 scores. 
The accurate cross-lingual label projection with an average of 82.1 F1 scores of 39 languages
guarantees the positive influence of our method to avoid excessive noise interference. 

\begin{figure}[t]
\begin{center}
	\includegraphics[width=1.0\columnwidth]{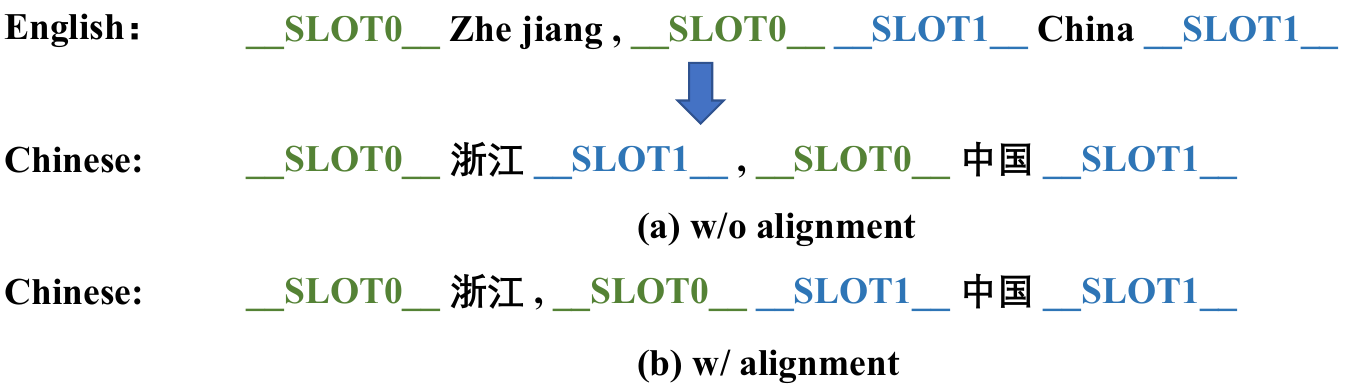}
	\caption{Comparison between the multilingual model w/o the alignment information and the counterpart w/ the alignment information in the training.}
	\label{translation_comparison}
\end{center}
\end{figure}

\paragraph{Example Study}
Table \ref{translation_comparison} lists a concrete example to compare our multilingual model with the baseline. In practice, we set the special token \texttt{\_\_SLOT{\{i\}}\_\_} as the boundary symbol $b_i$.
The entities are surrounded by ``\texttt{\_\_SLOT{\{i\}}\_\_}'' for translation, where ``\texttt{\_\_SLOT{\{i\}}\_\_}'' is used as the boundary symbol.
The positions of the boundary symbols ``\texttt{\_\_SLOT0\_\_}'' and ``\texttt{\_\_SLOT1\_\_}'' are misplaced during translation for the baseline model. In contrast, the multilingual model trained with alignment pairs accurately translates sentences and maintains the correct position of boundary symbols owing to the phrase-level alignment information.

\paragraph{Transfer for Distant Languages}
Compared to the transferability inaugurated by cross-lingual pretrained models, our method bridges the gap between the source language and distant target languages. The average F1 scores of similar and distant languages to English are denoted by Avg$_{sim}$ and Avg$_{dis}$. In Figure \ref{distant_languages}, Avg$_{sim}$ gains +5.1 points improvement while Avg$_{dis}$ outperforms XLM-R by a large margin +19.2 points. The NER model trained on the English corpus initialized by the pretrained model is easier to be extended to similar languages but is hard to be transferred to distant languages \cite{unsupervised_distant_languages}. Through cross-lingual entity projection, our method productively encourages knowledge transfer from the source language to distant languages contrasted with the baseline.
\begin{table}[t]
\centering
\resizebox{1.0\columnwidth}{!}{
\begin{tabular}{l|cccc|cccc}
\toprule
                        & de    & fr     & et   &Avg$_{sim}$ & ja & ta & zh &Avg$_{dis}$\\\midrule
XLM-R                   & 75.2  & 76.6   &72.2  &74.7           &21.0 &55.0     &27.7&34.6\\ 
+10K                    & 77.7  & 82.2   &76.1  &78.7           &42.4 &64.5     &48.9&51.3\\
+50K                    & 78.8  & 82.4   &76.8  &79.3           &44.6 &65.3     &49.4&53.1\\
+100K                   & 78.4  & 82.5   &77.0  &79.3           &45.0 &65.4     &49.3&53.2\\
+ALL                    & \textbf{79.6}  & \textbf{82.6}   &\textbf{77.2}  &\textbf{79.8} &\textbf{45.5} &\textbf{65.6}    &\textbf{50.3}&\textbf{53.8}\\
\bottomrule
\end{tabular}}
\caption{Evaluation results for similar and distant languages of the source language with different sizes of pseudo data. Avg$_{sim}$ and Avg$_{dis}$ represent the average F1 scores of similar languages and distant languages.}
\label{distant_languages}
\end{table}

\paragraph{Explanation for Entity Matching}
\begin{figure}[t]
\begin{center}
	\includegraphics[width=1.0\columnwidth]{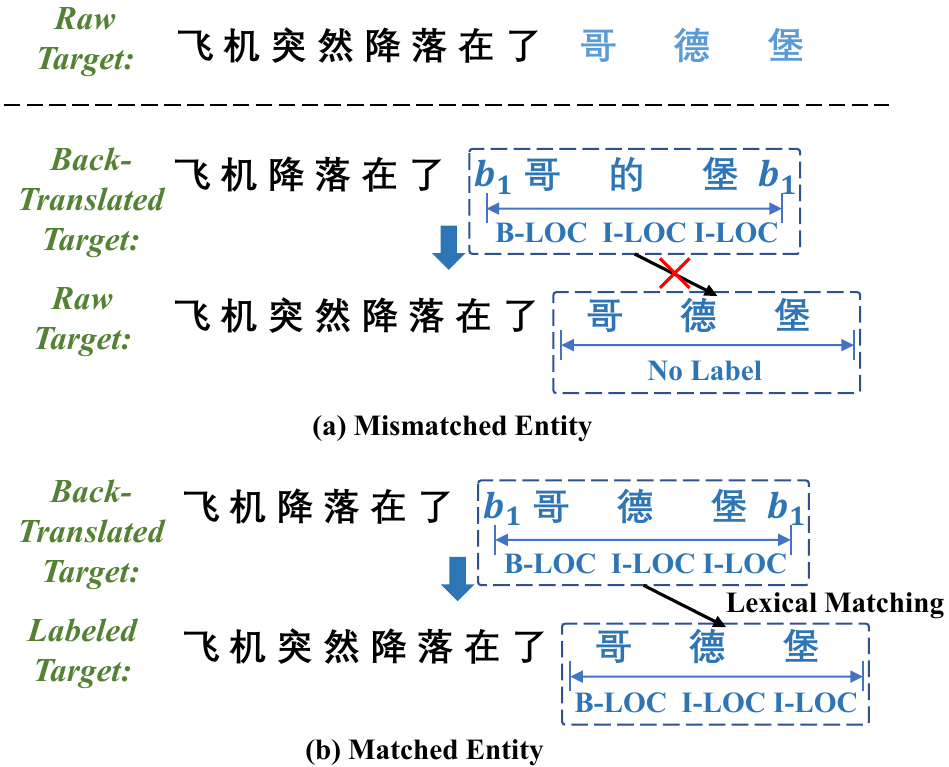}
	\caption{Entity matching includes (a) mismatched entity and (b) matched entity.}
	\label{entity_matching}
\end{center}
\vspace{-10pt}
\end{figure}
In Figure \ref{entity_matching}, we list two detailed examples of entity matching (a) mismatched entity and (b) matched entity. For the first example, ``哥的堡'' in the back-translated target is not mismatched to ``哥德堡'' in the raw target by the lexical matching, so the labels of ``哥的堡'' (\texttt{LOC}) can not be projected to the ``哥德堡''. For the second example, ``哥德堡'' in the back-translated target is the same as ``哥德堡'' in the raw sentence word by word, so we can obtain the labeled entity ``哥德堡'' (\texttt{LOC}) in the target sentence. The target sentences with missing entities are discarded, where the labels can not be projected to the entity like in the first example. Finally, we only need to select the 10\% sentences of all raw target sentences for the multilingual NER training to avoid extra noise and get state-of-the-art performance compared to previous baselines.

\section{Related Work}
\paragraph{Cross-lingual NER} Named entity recognition (NER) identifying the named entities into the predefined types has achieved huge progress in recent years \cite{conll_2003_ner,survey_ner,mrc_ner,modularized_interaction_ner,discontinuous_ner,zero_shot_ner,langauge_clustering_mner}. Cross-lingual NER model supporting multiple languages is a key component for various downstream natural language processing (NLP) tasks, including information retrieval \cite{information_retrieval_ner}, question answering \cite{question_answering_ner}, and co-reference resolution \cite{corefernence_resolution_ner}. The previous works can be classified into two categories including model-based transfer \cite{minimal_resources_ner,source_transfer_ner} and data-based transfer methods \cite{weak_supervision_ner,daga,mulda,melm}. The model-based transfer methods benefit from the state-of-the-art cross-lingual pretrained model \cite{mbert,xlmr} and the aligned cross-lingual word embeddings \cite{minimal_resources_ner}. \citet{unitrans} emphasizes that the model-based transfer and data-based transfer methods are complementary to each other.

\paragraph{Multilingual Translation} Inspired by the success of the neural machine translation \cite{nmt,transformer}, multilingual translation has attracted considerable attention due to
its capability to handle multiple languages in a shared single model \cite{ctl_mnmt,important_based_neuron_mnmt,distributionally_mnmt,competence_based_mnmt}. Previous works explicitly leverage the word-level or phrase-level extracted alignment information to improve performance. \cite{code_switching_nmt,csp,multilingual_agreement_mnmt}. Recently, massively multilingual models \cite{m2m,flores} are proposed, which all are trained on large sources of training data.
Motivated by previous works, we combine the phrase-level alignment pairs and the many-to-many multilingual model covering 40 languages to construct a labeled sequence translation system for the cross-lingual NER task in this work.

\section{Conclusion}
In this work, we propose a novel zero-shot cross-lingual NER framework with a multilingual labeled sequence translation model advised by multilingual corpora and phrase-level alignment pairs. The knowledge of the source NER model is effectively transferred to target languages by a round-trip translation and label projection. In this way, the multilingual translation model plays the role of the bridge to transfer knowledge from source languages to low-resource target languages. Experimental results evaluated on the CoNLL-5 and XTREME-40 benchmarks demonstrate the effectiveness of our method compared to the strong baselines.

\section{Limitations}
The total number of languages in our multilingual labeled sequence translation was limited owing to the data availability of cross-lingual NER.
Once NER datasets of more languages are available, we can train a stronger multilingual translation model to further enhance the overall performance.
In future work, our method can be scaled up to hundreds of languages to meet the needs of practical industrial scenarios. 

\section*{Acknowledgments}
This work was supported in part by the National Natural Science Foundation of China (Grant Nos. 62276017, U1636211, 61672081), the 2022 Tencent Big Travel Rhino-Bird Special Research Program, and the Fund of the State Key Laboratory of Software Development Environment (Grant No. SKLSDE-2021ZX-18).

\bibliography{anthology,custom}
\bibliographystyle{acl_natbib}

\end{CJK*}

\end{document}